%% file: main.tex
\documentclass[australian,english]{article}
\usepackage[latin9]{inputenc}
\usepackage{geometry}
\usepackage{array}
\usepackage{multirow}
\usepackage{amsmath}
\usepackage{amsthm}
\usepackage{amssymb}
\usepackage{graphicx}

\makeatletter

\providecommand{\tabularnewline}{\\}

\usepackage{microtype}
\usepackage{graphicx}
\usepackage{subfigure}
\usepackage{booktabs} 
\usepackage{hyperref}
\usepackage{lineno}

\date{}

\makeatother

\usepackage{babel}
\begin{document}
\title{Neurocoder: Learning General-Purpose Computation Using Stored Neural
Programs}
\author{Hung Le and Svetha Venkatesh\\
Applied AI Institute, Deakin University, Geelong, Australia\\
\texttt{\{thai.le,svetha.venkatesh\}@deakin.edu.au}}
\maketitle
\begin{abstract}
Artificial Neural Networks are uniquely adroit at machine learning
by processing data through a network of artificial neurons. The inter-neuronal
connection weights represent the learnt Neural Program that instructs
the network on how to compute the data. However, without an external
memory to store Neural Programs, they are restricted to only one,
overwriting learnt programs when trained on new data. This is functionally
equivalent to a special-purpose computer. Here we design Neurocoder,
an entirely new class of general-purpose conditional computational
machines in which the neural network ``codes'' itself in a data-responsive
way by composing relevant programs from a set of shareable, modular
programs. This can be considered analogous to building Lego structures
from simple Lego bricks. Notably, our bricks change their shape through
learning. External memory is used to create, store and retrieve modular
programs. Like today's stored-program computers, Neurocoder can now
access diverse programs to process different data. Unlike manually
crafted computer programs, Neurocoder creates programs through training.
Integrating Neurocoder into current neural architectures, we demonstrate
new capacity to learn modular programs, handle severe pattern shifts
and remember old programs as new ones are learnt, and show substantial
performance improvement in solving object recognition, playing video
games and continual learning tasks. Such integration with Neurocoder
increases the computation capability of any current neural network
and endows it with entirely new capacity to reuse simple programs
to build complex ones. For the first time a Neural Program is treated
as a datum in memory, paving the ways for modular, recursive and procedural
neural programming.
\end{abstract}
\global\long\def\softmax{\mathrm{softmax}}%

\global\long\def\sigmoid{\mathrm{sigmoid}}%

\global\long\def\softplus{\mathrm{softplus}}%

\section{Introduction}

\input{intro.tex}

\section{System}

\input{system.tex}

\section{Methods}

\input{method.tex}

\section{Results\label{sec:Results-1}}

\input{exp.tex}

\section{Discussion}

\input{discuss.tex}

\bibliographystyle{plain}
\bibliography{nsa}

\section*{\cleardoublepage Appendix}

\renewcommand\thesubsection{\Alph{subsection}}

\input{appendix.tex}

\end{document}

%% file: intro.tex
\begin{figure}
\begin{centering}
\includegraphics[width=0.8\textwidth]{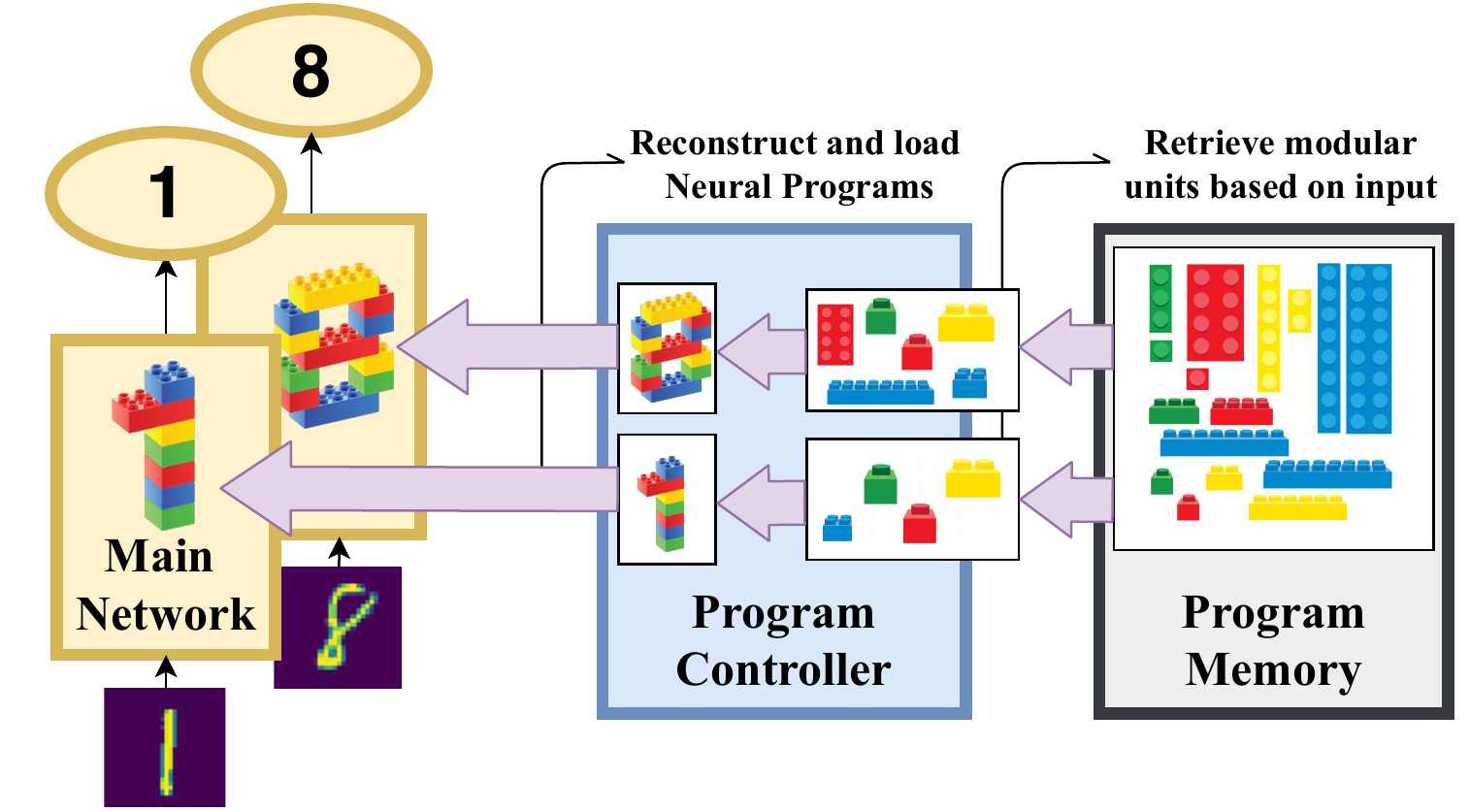}
\par\end{centering}
\caption{Overview structure of Neurocoder. Main Network processes inputs to
produce outputs. Program Memory stores modular units. Program Controller
reads modular units from the Program Memory, composing Neural Programs
for the Main Network in a data-driven manner. Each Neural Program
is designed for specific input. Intuitively, it is analogous to building
Lego structures corresponding to inputs from basic Lego bricks. \label{fig:overview}}
\end{figure}
From its inception in 1943 until recently, the fundamental architectures
of Artificial Neural Networks remained largely unchanged - a program
is executed by passing data through a network of artificial neurons
whose inter-neuronal connection weights are learnt through training
with data. These inter-neuronal connection weights, or Neural Programs,
correspond to a program in modern computers \cite{schmidhuber1990making}.
Memory Augmented Neural Networks (MANN) are an innovative solution
allowing networks to access external memory for manipulating data
\cite{graves2014neural,graves2016hybrid}. But they were still unable
to store Neural Programs in such external memory, and this severely
limits machine learning. Storing inter-neuronal connection weights
only in their network does not permit modular separation of Neural
programs and is analogous to a computer with one fixed program. Recent
works introduce \emph{conditional computation, }adjusting or activating
parts of a network in an input-dependent manner \cite{cogprints1380,schmidhuber1992learning,bengio2013estimating,ha2016hypernetworks},
but networks remain monolithic. Current networks forget when retrained,
old inter-neuronal connection weights are merged with new ones or
erased. 

The brain is modular, not a monolithic system \cite{edelman1978mindful,eccles1981modular}.
Neuroscience research indicates that the brain is divided into functional
modules \cite{hubel1988eye,edelman1993neural,frackowiak2004human}.
If the neural program for each module is kept in separate networks,
networks proliferate. Modular neural networks combine the output of
multiple expert networks, but as the experts grow, the networks grow
drastically \cite{jacobs1991adaptive,happel1994design,shazeer2017outrageously,rosenbaum2018routing}.
This requires huge computational storage and introduces redundancy
as these experts do not share common basic programs.

\emph{A pathway out of this bind is to keep such basic programs in
memory and combine them as required.} This brings neural networks
towards modern general-purpose computers that use the stored-program
principle \cite{turing1936,von1993first} to efficiently access reusable
programs in external memory. Here we show how Neurocoder, a new neural
framework, introduces an entirely new class of general-purpose conditional
computation machines in which an entire main neural network can be
``coded'' in an input-dependent manner. Efficient decomposition
of Neural Programs creates shareable modular components that can reconstruct
the whole program space. These components change their ``shapes''
based on training and are stored in an external Program Memory. Then,
in a data-responsive way, a Program Controller retrieves relevant
modular components to reconstruct the Neural Program. The process
is analogous to shape-shifting Lego bricks that can be reused to build
unlimited shapes and structures (See Fig. \ref{fig:overview}). 

Using modular components vastly increases the learning capacity of
the neural network by allowing re-utilisation of parameters, effectively
curbing network growth as programs increase. The construction of modular
components and the input-specific reconstruction of Neural Programs
from the components is learnt through training via traditional backpropagation
\cite{rumelhart1986learning} as the architecture is end-to-end differentiable.

%% file: system.tex
\begin{figure}
\begin{centering}
\includegraphics[width=1\textwidth]{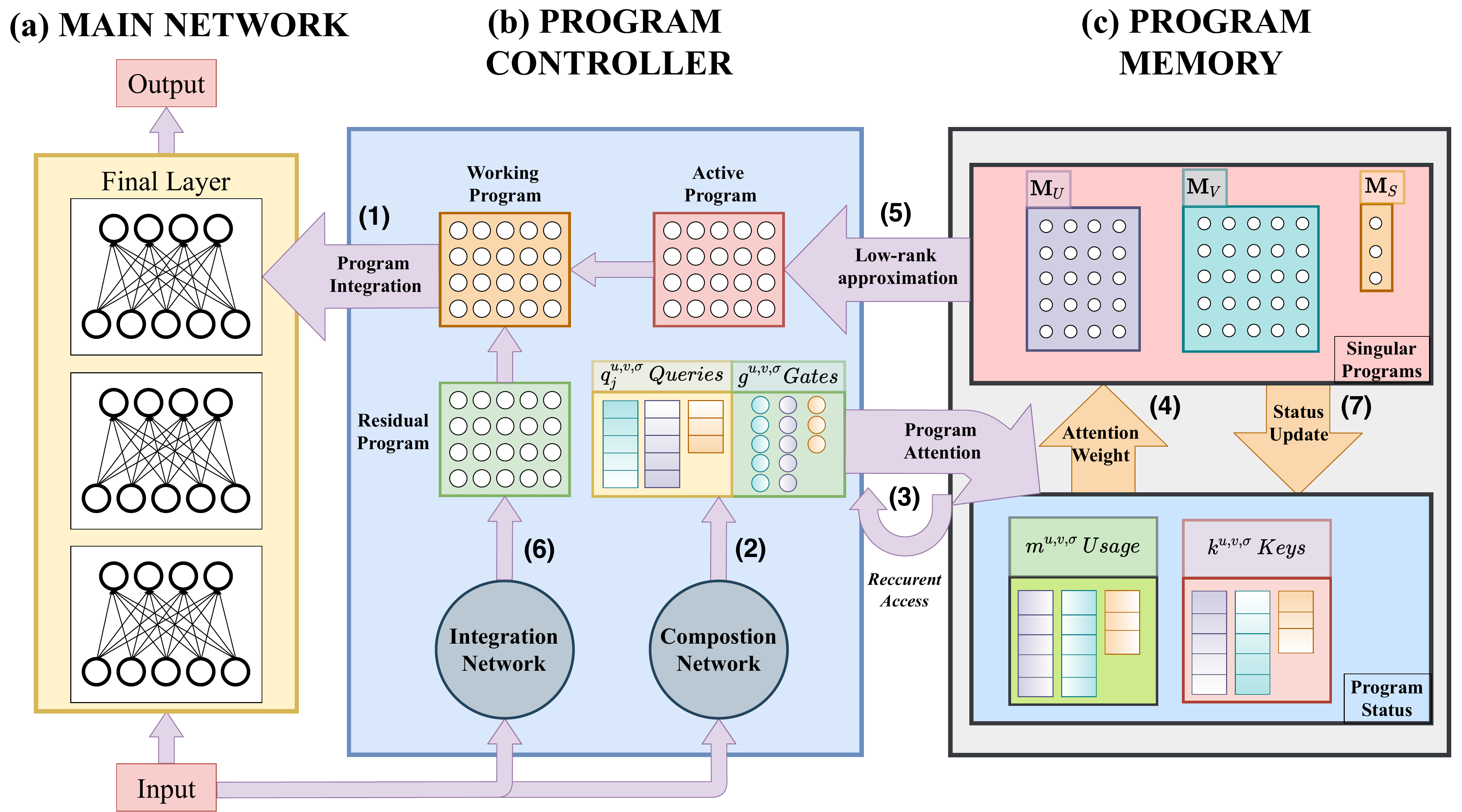}
\par\end{centering}
\caption{Neurocoder \textbf{(a)} The \emph{Main Network} uses a \emph{working
program} to compute the output for the input. Here only the final
layer of the Main Network is adaptively loaded with the working program
\textbf{(}\textbf{\emph{1}}\textbf{)}. Other layers use traditional
Neural Programs as connection weights (fixed-after-training). \textbf{(b)}
The Program Controller's \emph{composition} \emph{network} controls
access to the Program Memory, emitting queries and interpolating gate
control signals in response to the input \textbf{(}\textbf{\emph{2}}\textbf{)}.
It then performs recurrent multi-head program attention to the Program
Status \textbf{\emph{(3)}}, triggering attention weights to the Singular
Programs \textbf{\emph{(4)}}. The attended Singular Programs form
an \emph{active program} using low-rank approximation\emph{ }\textbf{\emph{(5)}}.
This active program is then used to derive the \emph{working program}
from a \emph{residual program} produced by the Program Controller's
\emph{integration network} \textbf{\emph{(}}\textbf{6)}. \textbf{(c)
}The Program Memory stores the representations (singular programs)
required to reconstruct the active program to be used by the Program
Controller. Access is controlled through the Program Status including
keys ($k$), and slot usage ($m$) that are updated during the training
and computation \textbf{\emph{(7)}}. \label{fig:Neural-Stored-program-Architectu}}
\end{figure}
A Neurocoder is a neural network (Main Network) coupled to an external
Program Memory through a Program Controller. The \emph{working program}
of the Main Network processes the input data to produce the output.
This working program is ``coded'' by the Program Controller by creating
an input-dependent \emph{active program }from the Program Memory (Fig.
\ref{fig:Neural-Stored-program-Architectu}).

\subsubsection*{Neurocoder stores Singular Value Decomposition of Neural Programs
in Program Memory}

The Neural Program needs to be stored efficiently in Program Memory.
This is challenging as there may be millions of inter-neuronal connection
weights, thus storing them directly (\cite{Le2020Neural}) is grossly
inefficient. Instead, the Neurocoder forms the basis of a subspace
spanned by Neural Programs and stores the singular values and vectors
of this subspace in memory slots of the Program Memory (hereafter
referred to as\emph{ singular program}s). Based on the input, relevant
singular programs are retrieved, a new program is reconstructed and
then loaded in the Main Network to process the input. This representational
choice significantly reduces the number of stored elements and allows
each singular program to effectively represent a unitary function
of the active program.

The \emph{active program} matrix $\mathbf{P}$ can be composed by
standard low-rank approximation as
\begin{equation}
\mathbf{P}=\mathbf{USV^{T}}\label{eq:Pusv}
\end{equation}
where $\mathbf{U}$ and $\mathbf{V}$ are matrices of the left and
right singular vectors, and $\mathbf{S}$ the matrix of singular values.
The Program Memory is crafted as three \emph{singular program memories}
$\left\{ \mathbf{M}_{U},\mathbf{M}_{V},\mathbf{M}_{S}\right\} $ to
store and retrieve these components. \emph{The process ``codes''
the active program using singular programs from }$\left\{ \mathbf{M}_{U},\mathbf{M}_{V},\mathbf{M}_{S}\right\} $. 

The Program Memory also maintains the status for each singular program
in terms of access and usage. To access a singular program, \emph{program
keys} ($k$) are used. These keys are low-dimensional vectors that
represent the singular program function and computed by a neural network
that effectively compresses the content of memory slots.\emph{ }The
\emph{program usage }($m$) measures memory utilisation, recording
how much a memory slot is used in constructing a program. The components
of the Program Memory are summarised in Fig. \ref{fig:Neural-Stored-program-Architectu}
(c).

\paragraph*{Recurrent multi-head program attention mechanisms for program storage
and retrieval}

Neural networks use the concept of \emph{differentiable attention
}to access memory \cite{graves2014neural,bahdanau2014neural,icml2020_1397}.
This defines a weighting distribution over the memory slots essentially
weighting the degree to which each memory slot participates in a read
or write operation. This is unlike conventional computers that use
a unique address to access a single memory slot.

Here we use two kinds of attention. First is \emph{content-based attention}
\cite{graves2014neural,graves2016hybrid} to ensure that the singular
program is selected based on its functionality and the data input.
This is achieved by producing a query vector based on the input and
comparing it to the program keys ($k$) using cosine similarity. Higher
cosine similarity scores indicate higher attention weights to the
singular programs associated with those program keys. Second, to encourage
better memory \foreignlanguage{australian}{utilisation}, higher attention
weights are assigned to slots with lower program usage ($m$) through
\emph{usage-based attention} \cite{graves2016hybrid,santoro2016meta}.
The attention weights from the two schemas are then combined using
interpolating gates to compose the final attention weights to the
Program Memory.

We adapt multi-head attention \cite{graves2014neural,vaswani2017attention}
that applies multiple attentions in parallel to retrieve $H$ singular
components. Besides, we introduce a recurrent attention mechanism,
in which multi-head access is performed recurrently in $J$ steps.
The $j$-th set of $H$ retrieved components is conditioned on the
previous ones. This recurrent, multi-head attention allows the composition
network to attend to multiple memory slots recurrently, incrementally
searching for optimal components for building relevant active programs.

\subsubsection*{Neurocoder learns to ``code'' a relevant working program via training}

The structure of the Program Memory and the role of the Program Controller
facilitates the automatic construction of working programs via training.
The Program Controller controls memory access through its \emph{composition}
\emph{network} that creates the \emph{attention weight} defining how
to weight the singular programs. Applying the recurrent multi-head
attention described earlier, multiple singular programs are retrieved
to construct an active program (Eq. \ref{eq:Pusv}). Then the Program
Controller generates a \emph{residual program }using its \emph{integration}
\emph{network} to transform the active program into the working program
of the Main Network. This transformation enables creation of flexible
higher-rank working programs, which compensates for the low-rank coding
process. The structure of the Program Controller is illustrated in
Fig. \ref{fig:Neural-Stored-program-Architectu} (b).

The singular programs are trained to represent unitary functions necessary
for any computation whilst the composition and integration networks
are trained to compose the relevant programs from the singular programs
to compute the current input. The parameters of the networks, and
the stored singular programs are adjusted during end-to-end training.
Initially, all the parameters will be random, leading to creation
of representations and working programs that produce huge training
loss. However, as training proceeds, these parameters get adjusted
and the network gets trained. Training is traditional and minimises
the total loss using gradient descent as

\begin{equation}
\mathcal{L}=\mathcal{L}_{task}+a\mathcal{L}_{o}\label{eq:final_loss}
\end{equation}
 where $\mathcal{L}_{task}$ represents the supervised training loss
and $\mathcal{L}_{o}$ represents a term weighted by a hyper-parameter
$a$ to enforce orthogonality of the singular vectors. 

\subsubsection*{Neurocoder can be integrated with any neural network}

The Main Network of the Neurocoder can be any current neural network.
One or more layers of the neural network can be replaced in this way
and re-coded with working programs from the Neurocoder in a data-responsive
way (Fig. \ref{fig:Neural-Stored-program-Architectu} (a)). Here we
show how Neurocoders re-code programs for single or all layers of
diverse networks like Multi-layer Perceptron, Convolutional Neural
Networks and Reinforcement Learning architectures. It is significant
that this can be done by plugging Neurocoder into current architectures
without modification of training paradigms or major addition to the
number of parameters. \emph{Thus any state-of-the-art neural architecture
plus Neurocoder can re-code itself from a suite of programs.}

%% file: method.tex
\subsubsection*{Program Coding as Low-rank Approximation \label{sec:Program-Construction-with}}

The Program Memory stores singular programs in three singular program
memories $\left\{ \mathbf{M}_{U},\mathbf{M}_{V},\mathbf{M}_{S}\right\} $.
At some time $t$, we compose the active program $\mathbf{P}_{t}$
by low-rank approximation as follows,
\begin{align}
\mathbf{P}_{t} & =\mathbf{US}\mathbf{V}^{T}\\
 & =\sum_{n}^{r_{m}}\sigma_{tn}u_{tn}v_{tn}^{\top}
\end{align}
where $r_{m}$ is the total number of components we want to retrieve,
$\left\{ \sigma_{tn}\right\} _{n=1}^{r_{m}}$ the singular values,
$\left\{ u_{tn}\right\} _{n=1}^{r_{m}}$ and $\left\{ v_{tn}\right\} _{n=1}^{r_{m}}$
the singular vectors of $\mathbf{S}$, $\mathbf{U}$, and $\mathbf{V}$,
respectively. 

By limiting $r_{m}$, we put a constrain on the rank of the active
program. To enforce orthogonality of the singular vectors, we minimise
the orthogonal loss

\begin{equation}
\mathcal{L}_{o}=\mathbf{M}_{U}\mathbf{M}_{U}^{\top}-\mathbf{I}+\mathbf{M}_{V}\mathbf{M}_{V}^{\top}-\mathbf{I}
\end{equation}

Since the active program is dynamically composed at time $t$ for
the computation of input $x_{t}$, it resembles fast-weights in neural
networks \cite{cogprints1380}. Unlike fast-weights, the working program
consists of singular programs stored in Program Memory representing
the stored-program principle \cite{turing1936,von1993first}. It differs
from an earlier attempt to implement a memory for programs, the Neural
Stored-program Memory \cite{Le2020Neural} which stores each item
as a working program itself. It extends the concept of slot-based
neural memory \cite{graves2014neural,graves2016hybrid,le2018variational,le2018learning}
to storing neural programs as data. 

We now use attention weights ($w_{tin}^{u}$, $w_{tin}^{v}$, $w_{tin}^{\sigma}$
jointly denoted as $w_{tin}^{u,v,\sigma}$) to each slot of the singular
program memories $\mathbf{M}_{U}$, $\mathbf{M}_{V}$ and $\mathbf{M}_{S}$
to read each singular vector as

\begin{equation}
u_{tn}=\sum_{i=1}^{P_{u}}w_{tin}^{u}\mathbf{M}_{U}\left(i\right)
\end{equation}

\begin{equation}
v_{tn}=\sum_{i=1}^{P_{v}}w_{tin}^{v}\mathbf{M}_{V}\left(i\right)
\end{equation}
For the singular values, we enforce $\sigma_{t1}>\sigma_{t2}>...>\sigma_{tr_{m}}>0$
by using

\begin{equation}
\sigma_{tn}=\begin{cases}
\softplus\left(\sum_{i=1}^{P_{s}}w_{tin}^{\sigma}\mathbf{M}_{S}\left(i\right)\right) & n=r_{m}\\
\sigma_{tn+1}+\softplus\left(\sum_{i=1}^{P_{s}}w_{tin}^{\sigma}\mathbf{M}_{S}\left(i\right)\right) & n<r_{m}
\end{cases}
\end{equation}
Here, $P_{u}$, $P_{v}$ and $P_{s}$ are the number of memory slots
of $\mathbf{M}_{U}$, $\mathbf{M}_{V}$ and $\mathbf{M}_{S}$, respectively.
For simplicity, in this paper, we set $P=P_{u}=P_{v}=P_{s}$ as the
number of memory slots of the Program Memory. The attention weights
$w_{tin}^{u,v,\sigma}$, shorten form for $w_{tijh}^{u,v,\sigma}$,
are determined by program memory attention mechanisms, which will
be discussed in the upcoming sections.

\begin{figure}
\begin{centering}
\includegraphics[width=1\textwidth]{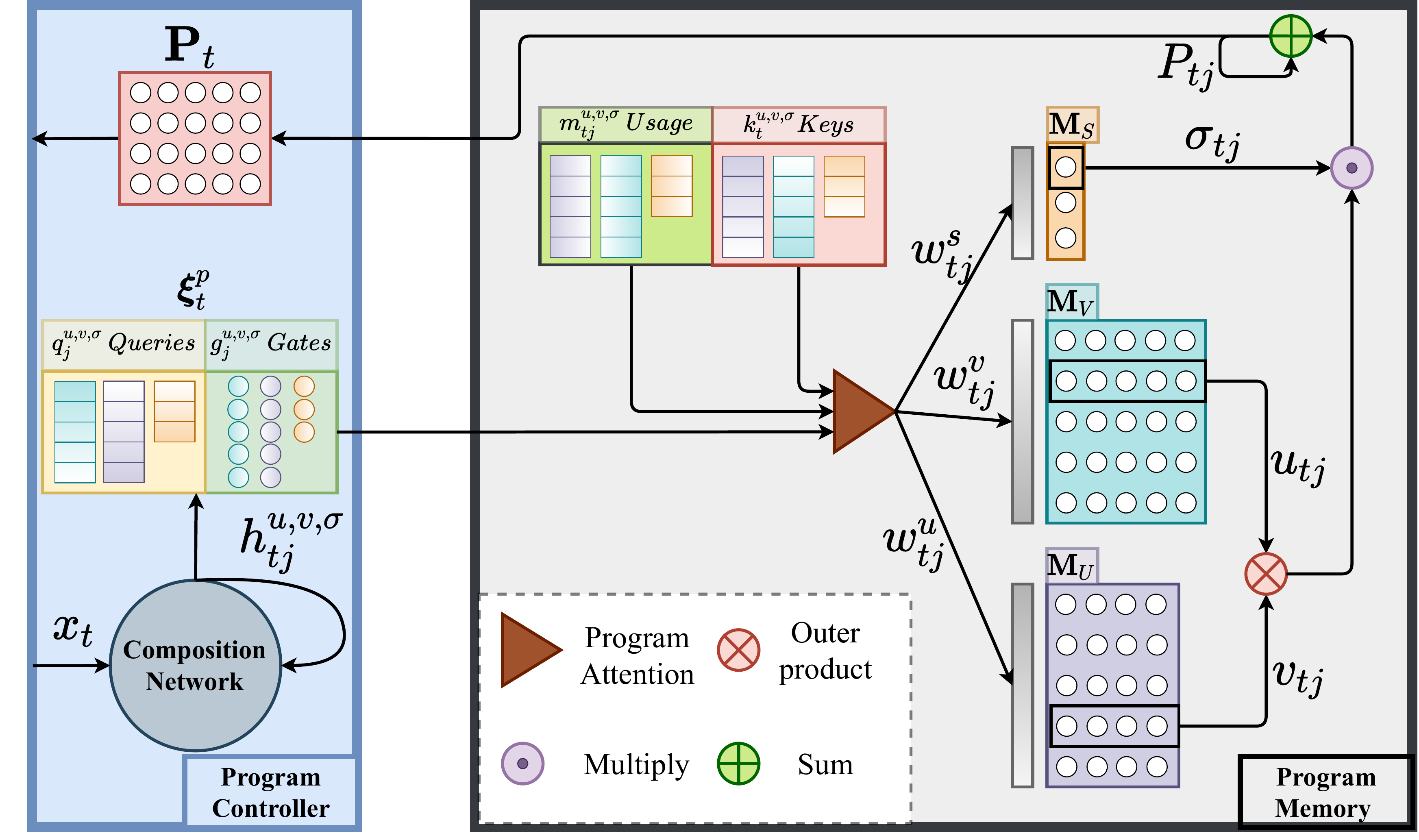}
\par\end{centering}
\caption{Active program coding. The Program Controller uses the composition
network (a recurrent neural network) to process the input \emph{$x_{t}$
}and generate composition signal $\boldsymbol{\xi}_{t}^{p}$, which
is composed of the queries ($q$) and the interpolating gates ($g$).
The similarity of the query to program memory keys ($k$) is then
computed together with the memory usage ($m$) from which attention
weights for the Program Memory are derived. The active program $\mathbf{P}_{t}$
is then ``coded'' through low-rank approximation using the $j$-th
component accessed by recurrent attentions. For simplicity, one attention
head is shown ($H=1$). \label{fig:coding}}
\end{figure}

\subsubsection*{Recurrent Access to the Program Memory via the composition network}

To perform program attention, the Program Controller employs a composition
network (denoted as $f_{\theta^{u,v,\sigma}}$), which takes the current
input $x_{t}$ and produce \emph{program composition control signals}
($\boldsymbol{\xi}_{t}^{p}$). If $f_{\theta^{u,v,\sigma}}$ performs
all attentions concurrently via multi-head attention (as in \cite{graves2014neural,vaswani2017attention}),
it may lead to program collapse \cite{Le2020Neural}. To have a better
control of the component formation and alleviate program collapse,
we propose to recurrently attend to the program memory. To this end,
we implement $f_{\theta^{u,v,\sigma}}$ as a recurrent neural network
(LSTM \cite{hochreiter1997long}) and let it access the program memory
$J$ times, resulting in $\boldsymbol{\xi}_{t}^{p}=\left\{ \boldsymbol{\xi}_{tj}^{p}\right\} _{j=1}^{J}$.
At access step $j$, the recurrent network updates its hidden states
and generates $\boldsymbol{\xi}_{tj}^{p}$ using recurrent dynamics
as

\begin{equation}
\boldsymbol{\xi}_{tj}^{p},h_{j}^{u,v,\sigma}=f_{\theta^{u,v,\sigma}}\left(x_{t},h_{j-1}^{u,v,\sigma}\right)\label{eq:rnn_q}
\end{equation}
where $h_{0}^{u,v,\sigma}$ is initialized as zeros and $\boldsymbol{\xi}_{tj}^{p}$
is the\emph{ }program composition control signal at step $j$ that
depends on both on the input data $x_{t}$ and the the previous state
$h_{j-1}^{u,v,\sigma}$. Particularly, the control signal contains
the queries and the interpolation gates used to compute the program
attention ($w_{tin}^{u,v,\sigma}$): $\boldsymbol{\xi}_{tj}^{p}=\left\{ q_{tjh}^{u,v,\sigma},g_{tijh}^{u,v,\sigma}\right\} _{h=1}^{H}$.
Here, at each attention step, we perform multi-head attention with
$H$ as the number of attention heads or retrieved components and
thus, each $\boldsymbol{\xi}_{tj}^{p}$ consists of $H$ pairs of
queries and gates. Hence, the total number of retrieved components
$r_{m}=J\times H$.

\subsubsection*{Attending to Programs by ``Name''}

 Inspired by the content-based attention mechanism for data memory
\cite{graves2014neural}, we use the query to look for the singular
programs. In computer programming, to find the appropriate program
for some computation, we often refer to the program description or
at least the name of the program. Here, we create the ``name'' for
our neural programs by compressing the program content to a low-dimensional
key vector. As such, we employ a neural network ($f_{\varphi}$) to
compute the program memory keys as

\begin{equation}
k_{i}^{u,v,\sigma}=f_{\varphi^{u,v,\sigma}}\left(\mathbf{M}_{U,V,S}\left(i\right)\right)
\end{equation}
where $k^{u,v,\sigma}=\left\{ \left\{ k_{i}^{u}\in\mathbb{R}^{K}\right\} _{i=1}^{P_{u}},\left\{ k_{i}^{v}\in\mathbb{R}^{K}\right\} _{i=1}^{P_{v}},\left\{ k_{i}^{\sigma}\in\mathbb{R^{K}}\right\} _{i=1}^{P_{\sigma}}\right\} $.
Here, $f_{\varphi^{u,v,\sigma}}$ learns to compress each memory slot
of the singular program memories into a $K$-dimensional vector. As
the singular programs evolve, their keys get updated. In this paper,
we calculate the program keys after each learning iteration during
training. 

Finally the content-based program memory attention $c_{tijh}^{u,v,\sigma}=\left\{ c_{tijh}^{u},c_{tijh}^{v},c_{tijh}^{\sigma}\right\} $
is computed using cosine distance between the program keys $k_{i}^{u,v,\sigma}$
and the queries $q_{tjh}^{u,v,\sigma}$ as

\begin{equation}
c_{tijh}^{u,v,\sigma}=\softmax^{(i)}\left(\frac{q_{tjh}^{u,v,\sigma}\cdot k_{i}^{u,v,\sigma}}{||q_{tjh}^{u,v,\sigma}||\cdot||k_{i}^{u,v,\sigma}||}\right)\label{eq:d_p-1}
\end{equation}

\subsubsection*{Making Every Program Count}

Similarly to \cite{graves2016hybrid,santoro2016meta}, in addition
to the content-based attention, we employ a least-used reading strategy
to encourage the Program Controller to assign different singular programs
to different components. In particular, we calculate the memory usage
for each program slot across attentions as 
\begin{equation}
m_{tijh}^{u,v,\sigma}=\underset{\tilde{j}\leq j}{\max}\left(w_{ti\tilde{j}h}^{u,v,\sigma}\right)
\end{equation}
 where $m^{u,v,\sigma}=\left\{ m^{u}\in\mathbb{R}^{P_{u}\times1},m^{v}\in\mathbb{R}^{P_{v}\times1},m^{\sigma}\in\mathbb{R}^{P_{\sigma}\times1}\right\} $.
Since we want to consider only $l_{I}$ amongst $P$ memory slots
that have smallest usages, let $\hat{m}_{tjh}^{u,v,\sigma\left(l_{I}\right)}$
denote the value of the $l_{I}$-th smallest usage, then the least-used
attention is computed as

\begin{equation}
l_{tijh}^{u,v,\sigma}=\begin{cases}
\underset{i}{\max}\left(m_{tijh}^{u,v,\sigma}\right)-m_{tijh}^{u,v,\sigma} & ;m_{tijh}^{u,v,\sigma}\leq\hat{m}_{tjh}^{u,v,\sigma\left(l_{I}\right)}\\
0 & ;m_{tijh}^{u,v,\sigma}>\hat{m}_{tjh}^{u,v,\sigma\left(l_{I}\right)}
\end{cases}
\end{equation}
The final program memory attention is computed as

\begin{equation}
w_{tijh}^{u,v,\sigma}=\sigmoid\left(g_{tijh}^{u,v,\sigma}\right)c_{tijh}^{u,v,\sigma}+\left(1-\sigmoid\left(g_{tijh}^{u,v,\sigma}\right)\right)l_{tijh}^{u,v,\sigma}
\end{equation}
Since the usage record are computed along the memory accesses, the
multi-step Neurocoder utilises this attention mechanism better than
the single-step Neurocoder, leading to different attention behaviors
(see Sec. \ref{sec:Results-1}). The whole process of composing the
active program $\mathbf{P}_{t}$ is illustrated in Fig. \ref{fig:coding}.

\subsubsection*{Program Integration via the integration network \label{sec:Program-Utilization-for}}

Since the working program $\mathbf{P}_{t}$ only contains top $r_{m}$
principal components, it may be not flexible enough for sophisticated
computation. We propose to enhance $\mathbf{P}_{t}$ with a residual
program $\mathbf{R}$-- a traditional connection weight trained as
the integration network's parameters. The residual program represents
the sum of the remaining less important components. To this end, we
suppress $\mathbf{R}$ with a multiplier that is smaller than $\sigma_{tr_{m}}$--
the smallest singular value of the main components - resulting in
the integration formula

\begin{equation}
W_{t}=\mathbf{P}_{t}+w_{t}^{r}\sigma_{tr_{m}}\mathbf{R}\label{eq:util}
\end{equation}
where $w_{t}^{r}=\sigmoid\left(f_{\phi}\left(x_{t}\right)\right)$
is an adaptive gating value that controls the contribution of the
residual program. $f_{\phi}$ is the integration network in the Program
Controller and hence, in our implementation, the integration control
signal sent by the Program Controller is $\boldsymbol{\lambda}_{t}^{p}=\left\{ w_{t}^{r},\sigma_{tr_{m}}\right\} .$
We note that in our experiments, the program integration is sometimes
disabled ($W_{t}$ is directly set to $\mathbf{P}_{t}$) to eliminate
the effect of $\mathbf{R}$ or reduce the number of parameters. 

The working program $W_{t}$ is then used by the Main Network to execute
the input data $x_{t}$. For example, with linear classifier Main
Network, the execution is $y_{t}=x_{t}W_{t}$. Table \ref{tab:Neurocoder-parameters}
summarises the notations used for important parameters of Neurocoder. 

The Main Network can be any neural network in which one or more layers
of this network can be replaced by the Neurocoder. In our experiments,
we always apply Neurocoder to all layers of multi-layer perceptrons
(MLP) or just the final output layer of CNNs (LeNet, DenseNet, ResNet),
RNNs (GRU), and the policy/value networks of A3C. Other competitors
such as MOE, NSM and HyperNet are applied to the Main Networks in
the same manner.

%% file: exp.tex
\begin{figure}
\begin{centering}
\includegraphics[width=1\textwidth]{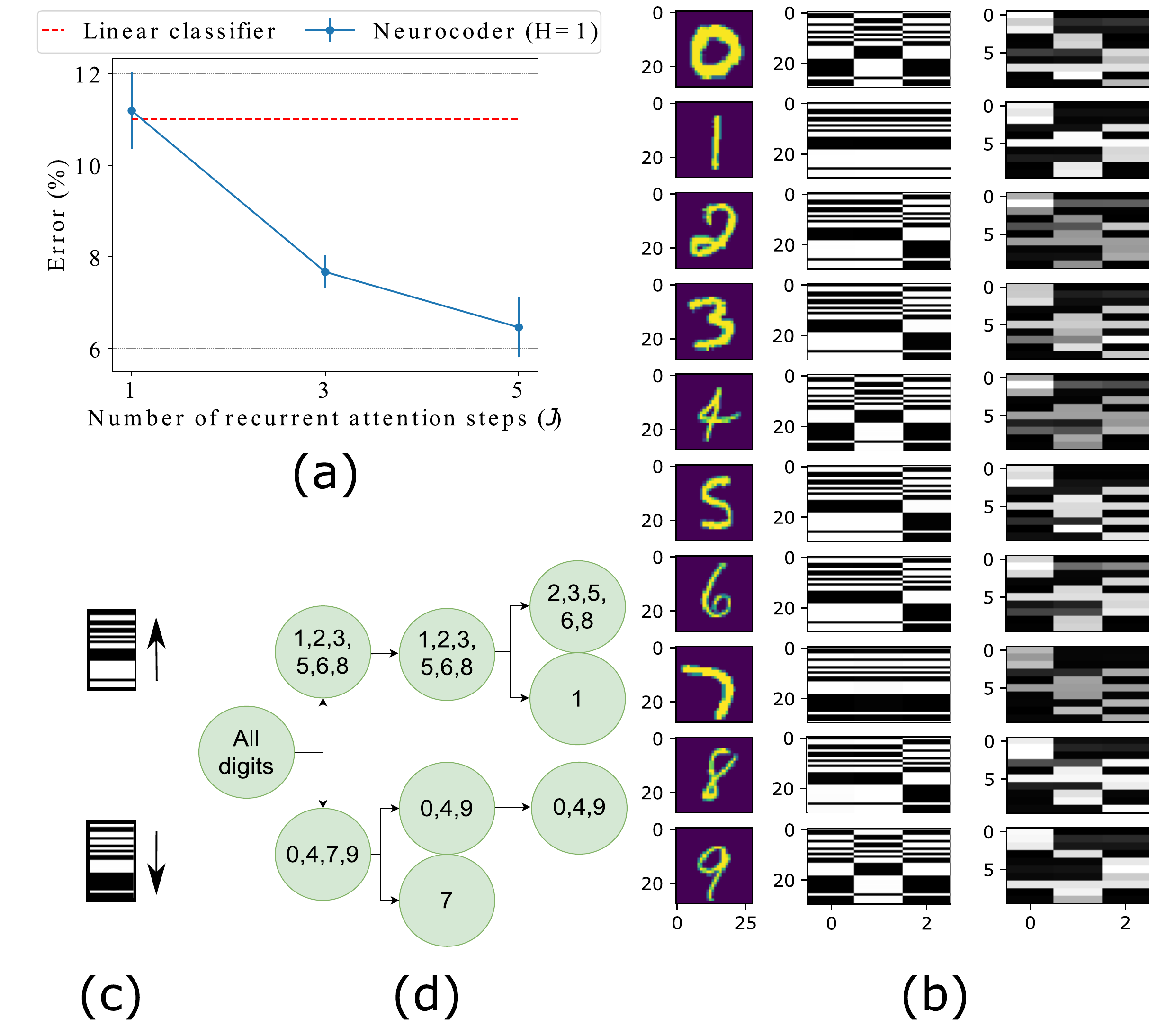}
\par\end{centering}
\caption{\textbf{(a)} MNIST test set classification error \emph{vs} the number
of steps ($J$) in Neurocoder (blue), compared with a linear classifier
(red). \textbf{(b) }\emph{1st column}: Digit images; \emph{Middle
column}: Single-step attention weights for $30$ slots in $\mathbf{M}_{U}$
(vertical axis) for first $3$ singular vectors (horizontal axis)
for each digit; \emph{Last column}: Multi-step attention weights for
$10$ slots in $\mathbf{M}_{U}$ (vertical axis) for first $3$ singular
vectors (horizontal axis). Multi-step attention is able to produce
far more diverse patterns with fewer slots - $10$ slots compared
to single-step $30$ slots. \textbf{(c)} Two attention patterns of
single-step Neurocoder. \textbf{(d)} The binary decision tree derived
from single-step Neurocoder's attention patterns. The two patterns
across components represent the decisions going up and down across
the binary tree. \label{fig:linear_mnist}}
\end{figure}
To demonstrate the flexibility of this framework, we consider different
learning paradigms: instance-based, sequential and continual learning.
We do not focus on breaking performance records by augmenting state-of-the-art
models with Neurocoder. Rather our inquiry is on re-coding layers
of diverse neural networks with the Neurocoder's programs and testing
on varied data types to demonstrate its intrinsic properties (details
of the following section are in Appendix).

\subsection{Instance-based learning - Object Recognition}

We tested Neurocoder on instance-based learning through classical
image classification tasks using MNIST \cite{lecun1998gradient} and
CIFAR \cite{krizhevsky2009learning} datasets. The first experiment
interpreted Neurocoder's behaviour in classifying digits into $10$
classes ($0-9$) using linear classifier Main Network. With equivalent
model size, Neurocoder using the novel recurrent attention surpasses
the performance of the linear classifier \cite{lecun1998gradient}
by up to $5\%$ (Fig. \ref{fig:linear_mnist} (a)). 

To differentiate the input, Neurocoder attends to different components
of the active program to guide the decision-making process. Fig. \ref{fig:linear_mnist}
(b) shows single-step and multi-step attention to the first $3$ singular
vectors for each digit across memory slots. Multi-step attention produces
richer patterns compared to single-step Neurocoder that manages only
$2$ attention weight patterns (Fig. \ref{fig:linear_mnist} (c)). 

Fig. \ref{fig:linear_mnist} (d) illustrates how Neurocoder performs
modular learning by showing the attention assignment for top $3$
singular vectors as a binary decision tree. Digits under the same
parental node share similar attention paths, and thereby similar active
programs. Some digits look unique (e.g. $7$) resulting in active
programs composed of unique attention paths, discriminating themselves
early in the decision tree. Some digits (e.g. $0$ and $9$) share
the same attention pattern for the first $3$ components and are thus
unclassifiable in the binary tree. They can only be distinguished
by considering more singular vectors.

We integrated Neurocoder with deep networks - \emph{$5$-layer LeNet
and $100$-layer DenseNet }- and tested on complex CIFAR datasets.
Neurocoder significantly outperformed the original Main Networks with
performance gain around $2-5\%$. Compared with recent conditional
computing models such as Mixture of Experts (MOE \cite{shazeer2017outrageously})
and Neural Stored-program Memory (NSM \cite{Le2020Neural}), Neurocoder
required a tenth of the number of parameters and performed better
by up to $7\%$ (see  Table \ref{tab:Best-accuracy-over}).

\begin{figure}
\begin{centering}
\includegraphics[width=1\textwidth]{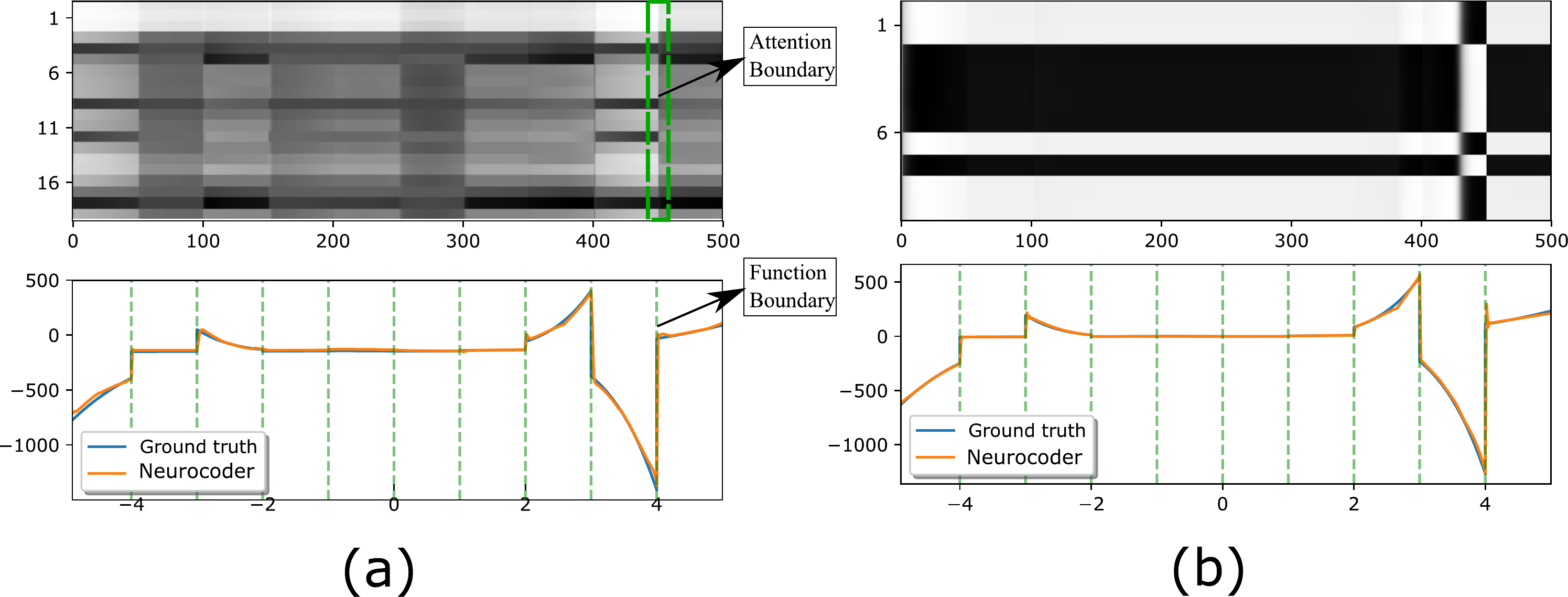}
\par\end{centering}
\caption{Visualisation for \textbf{(a)} multi-step ($J=5$, $20$ memory slots)
and \textbf{(b)} single-step ($J=1$, $10$ memory slots) cases showing
while processing a sequence of the polynomial auto-regression task.
The Neurocoder's attentions to $\mathbf{M}_{U}$ that form the first
component of the active program are shown over sequence timesteps
(\emph{upper}) with Neurocoder's $y_{t}$ prediction (orange) and
ground truth (blue) (\emph{lower}). The vertical dash green lines
separate polynomial chunks. Each chuck represents a local pattern,
and thus ideally requires a specific active program to compute the
input $x_{t}$. Although both predict well, only the multi-step Neurocoder
discovers the chunk boundaries, assigning program attention to the
first component in accordance with sequence changes. \label{fig:Attention-to-program}}
\end{figure}

\subsection{Sequential learning - Adaption to sequence changes and game playing
using reinforcement learning}

Recurrent neural networks (RNN) can learn from sequential data by
updating the hidden states of the networks. However, this does not
suffice when local patterns shift, as is often the case. We now demonstrate
that Neurocoder helps RNNs overcome this limitation by composing diverse
programs to handle sequence changes. 

\paragraph{Synthetic polynomial auto-regression }

We created a simple auto-regression task in which data points are
sampled from polynomial function chunks that change over time. The
Main Network is a strong \emph{RNN--Gated Recurrent Unit} (\emph{GRU}
\cite{cho2014learning}). We found that GRU integrated with a single-step
or multi-step Neurocoder learned much faster than its original version
and HyperNet counterparts \cite{ha2016hypernetworks} ( Fig. \ref{fig:Polynomial-autoregression:-mean}).

Visualising the first singular vector attention weights in $\mathbf{M}_{U}$,
we find that the multi-step attention Neurocoder changes its attention
following polynomial changes - it attends to the same singular program
when processing data from the same polynomial and alters attention
for data from a different polynomial (Fig. \ref{fig:Attention-to-program}(a)).
In contrast, the single-step Neurocoder only changes its attention
when there is a remarkable change in $y$-coordinate values (Fig.
\ref{fig:Attention-to-program}(b)). We \foreignlanguage{australian}{hypothesise}
that when recurrence is employed, usage-based attention takes effect,
stipulating better memory \foreignlanguage{australian}{utilisation}
and diverse attentions over timesteps. Although single-step Neurocoder
converges well, it did not discover the underlying structure of the
data, and thus underperformed the multi-step Neurocoder.

\paragraph{Atari game reinforcement learning}

We used reinforcement learning as a further testbed to show the ability
to adapt to environmental changes. We performed experiments on several
Atari 2600 games \cite{bellemare2013arcade} wherein the agent (or
Main Network) was implemented as the \emph{Asynchronous Advantage
Actor-Critic }(\emph{A3C} \cite{mnih2016asynchronous}). In the Atari
platform, agents are allowed to observe the screen snapshot of the
games and act to earn the highest score. We augmented the A3C by employing
Neurocoder's working programs for the actor and critic networks, aiming
to decompose the policy and value function into singular programs
that were selected depending on the game state.

\emph{Frostbite and Montezuma\textquoteright s Revenge}. These games
are known to be challenging for A3C and other algorithms \cite{mnih2016asynchronous}.
We trained A3C and HyperNet-based A3C for over $300$ million steps,
yet these models did not show any sign of learning, performing equivalently
to random agents. For such complicated environments with sparse rewards,
both the monolithic neural networks and the unstored fast-weights
fail to learn (almost zero scores). In contrast, Neurocoder enabled
A3C to achieve from $1,500$ to $3,000$ scores on these environments
(Fig. \ref{fig:atari}), confirming the importance of decomposing
a complex solution to smaller, simple stored programs. 

\subsection{Continual learning }

In continual learning, standard neural networks often suffer from
``catastrophic forgetting'' in which they cannot retain knowledge
acquired from old tasks upon learning new ones \cite{french1999catastrophic}.
Our Neurocoder offers natural mitigation of such catastrophic forgetting
in neural networks by storing task-dependent singular programs in
separated program memory slots. 

\paragraph{Split MNIST}

We first considered the split MNIST dataset--a standard continual
learning benchmark wherein the original MNIST was split into a $5$
$2$-way classification tasks, consecutively presented to a \emph{Multi-layer
Perceptron} Main Network (MLP). We followed the benchmarking as in
\cite{hsu2018re} in which various optimisers and state-of-the-art
continual learning methods were examined under incremental task and
domain scenarios. We measured the performance of the MLP versus Neurocoder
and NSM under each continual learning method. 

In both scenarios, Neurocoder was compatible with all continual leaning
methods, demonstrating superior performance over MLP and NSM with
performance gain between $1$ to $16\%$ (see  Table \ref{tab:Incremental-task-continual}
and \ref{tab:Incremental-domain-continual}). 

\paragraph{Split CIFAR}

We verified the scalability of Neurocoder to more challenging datasets.
We split CIFAR datasets as in the split MNIST, wherein $5$-task $2$-way
split CIFAR10 and a $20$-task $5$-way split CIFAR100 were created.
We integrate Neurocoder with \emph{ResNet} \cite{he2016deep}--a
very deep CNN architecture as the Main Network.

When we stressed the orthogonal loss ($a=10$) and used bigger program
memory ($100$ slots), Neurocoder improved ResNet classification by
$15\%$ and $10\%$ on CIFAR10 and CIFAR100, respectively. When we
integrated Neurocoder with Synaptic Intelligence (SI \cite{zenke2017continual}),
the performance was further improved, maintaining a stable performance
above $80\%$ accuracy for CIFAR10 and outperforming using SI alone
by $10\%$ for CIFAR100 (see  Fig. \ref{fig:Incremental-task-continual}).

%% file: discuss.tex
Our experiments demonstrate that Neurocoder is capable of re-coding
Neural Programs in distinctive neural networks, amplifying their capabilities
in diverse learning scenarios: instance-based, sequential and continual
learning. This consistently results in significant performance increase,
and further creates novel robustness to pattern shift and catastrophic
forgetting. This unprecedented ability for each architecture to re-code
itself is made possible without changing the way it is trained, or
majorly increasing the number of parameters it needs to learn.

The MNIST problem illustrates the reasoning process of Neurocoder
when classifying digit images wherein its singular program assignment
resembles a binary tree decision-making process - it shows how some
singular programs are shared, others are not. The polynomial auto-regression
problem highlights the importance of efficient memory utilisation
in re-constructing the working program enabling discovery of hidden
structures in sequential data. Training our framework with reinforcement
learning, we enable neural agents to solve complex games wherein traditional
methods fail or learn slowly. Finally, continual learning problems
show that Neurocoder mitigates catastrophic forgetting efficiently
under different learning settings/algorithms. 

Our solution offers a single framework that is scalable and adaptable
to various problems and learning paradigms. Unlike previous attempts
to employ a bank of separate big programs \cite{jacobs1991adaptive,shazeer2017outrageously,Le2020Neural},
Neurocoder maintains only shareable, smaller components that can reconstruct
the whole program space, thereby heavily utilising the parameters
and preventing the model from proliferating. We can further extend
Neurocoder's ability by allowing a growing Program Memory, in which
the model decides to add or erase memory slots as the number of data
patterns grows or shrinks beyond the current program space's capacity.
Such a system represents a more flexible general-purpose computer
that can dynamically allocate computing resources by itself without
human pre-specification.

%% file: appendix.tex
\subsubsection*{Instance-based learning experiments}

\paragraph*{\emph{Image classification-linear Main Network}}

We used the standard training and testing set of MNIST dataset. To
train the models, we used the standard SGD with a batch size of $32$.
Each MNIST image was flattened to a $768$-dimensional vector, which
requires a linear classifier of $7,680$ parameters to categorise
the inputs into $10$ classes. For Neurocoder, we used Program Memory
with $P=6$ and $K=2$. The Program Controller's composition network
was an LSTM with a hidden size of 8. We controlled the number of parameters
of Neurocoder, which included parameters for the Program Memory and
the Program Controller by reducing the input dimension using random
projection $z_{t}=x_{t}U$ with $U\in\mathbb{R}^{768\times200}$ initialised
randomly and fixed during the training. We also excluded the program
integration to eliminate the effect of the residual program $\mathbf{R}$.
Given the flattened image input $x_{t}$, Neurocoder generated the
active program $\mathbf{P}_{t}$, predicting the class of the input
as $y_{t}=argmax\left(x_{t}\mathbf{P}_{t}\right)$. The performance
of the linear classifier was imported from \cite{lecun1998gradient}
and confirmed by our own implementation.

\paragraph*{\emph{Image classification-deep Main Network}}

We used the standard training and testing sets of CIFAR datasets.
We use Adam optimiser with a batch size of 128. \emph{The deep Main
Networks were adopted from the original papers, resulting in $3$-layer
MLP, $5$-layer LeNet} \cite{lecun1998gradient}\emph{ and $100$-layer
DenseNet} \cite{huang2017densely}. The other baselines for this task
included a recent Mixture of Experts (MOE \cite{shazeer2017outrageously})
and the Neural Stored-program Memory (NSM \cite{Le2020Neural}). For
this case, we employed the program integration with the residual program
$\mathbf{R}$. The Main Network's hyper-parameters were fixed and
we only tuned the hyper-parameters of Neurocoder, MOE and NSM. We
report details of hyper-parameters in  Table \ref{tab:param} and
\ref{tab:param-1}. 

\subsubsection*{Sequential learning experiments}

\paragraph*{\emph{Synthetic polynomial auto-regression}}

A sequence was divided into $n_{pa}$ chunks, each of which associated
with a randomly generated polynomial. The degree and coefficients
of each polynomial were sampled from $U\sim\left[2,10\right]$ and
$U\sim\left[-1,1\right]$, respectively. Each sequence started from
$x_{1}=-5$ and ended with $x_{T}=5$, equally divided into $n_{pa}$
chunks. Each chunk contained several consecutive points $\left(x_{t},y_{t}\right)$
from the corresponding polynomial, representing a local transformation
from the input to the output. Given previous points $\left(x_{<t},y_{<t}\right)$
and the current $x$-coordinate $x_{t}$, the task was to predict
the current $y$-coordinate $y_{t}$. To be specific, at each timestep,
the Main Network GRU was fed with $\left(x_{t},y_{t-1}\right)$ and
trained to predict $y_{t}$ by minimizing the mean square error $1/T\sum_{t=1}^{T}\left(\hat{y_{t}}-y_{t}\right)^{2}$
where $y_{0}=0$, $\hat{y_{t}}$ is the prediction of the network
and $y_{t}$ the ground truth. 

We augmented \emph{GRU by applying Neurocoder} and HyperNet \cite{ha2016hypernetworks}
\emph{to the output layer of the GRU}. Here, the HyperNet baseline
generated adaptive scales for the output weight of the GRU. We trained
the networks with Adam optimiser with a batch size of $128$. To balance
the model size, we used GRU's hidden size of $32$, $28$, $16$ and
$8$ for the original Main Network, HyperNet, single-step and multi-step
Neurocoder, respectively. We also excluded program integration phase
in Neurocoders to keep the model size equivalent to or smaller than
that of the Main Network. We report details of hyper-parameters for
GRU, HyperNet and Neurocoder in  Table \ref{tab:param} and \ref{tab:param-1}. 

\begin{figure}
\begin{centering}
\includegraphics[width=1\textwidth]{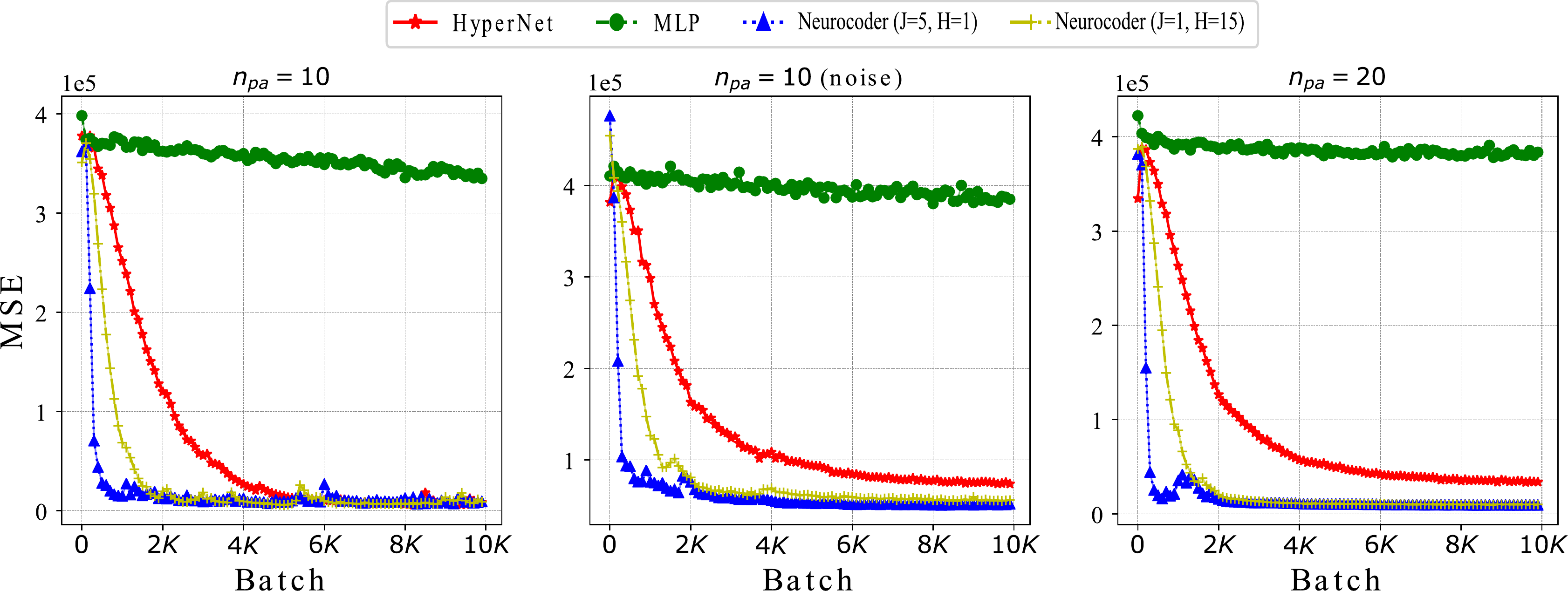}
\par\end{centering}
\caption{Polynomial auto-regression: mean square error (MSE) over training
iterations with a batch size of 128 comparing HyperNet (red), MLP
(green), multi-step (blue) and single-step (yellow) attention Neurocoders.
All baselines use GRU as the Main Network. The learning curves are
taken average over 5 runs. \label{fig:Polynomial-autoregression:-mean}}
\end{figure}

We compared two configurations of Neurocoder - single-step, multi-head
($J=1$, $H=15$) and multi-step, single-head ($J=5$, $H=1$)- against
the original GRU with output layer made by MLP and HyperNet--a weight-adaptive
neural network \cite{ha2016hypernetworks}. We found that MLP failed
to learn and converge within $10,000$ learning iterations. In contrast,
both Neurocoders learn and converge, in as little as only $2,000$
iterations with the multi-step Neurocoder. HyperNet converged much
slower than Neurocoders and could not minimize the predictive error
as well as Neurocoders when Gaussian noise is added or the number
of polynomials ($n_{pa}$) is doubled (see  Fig. \ref{fig:Polynomial-autoregression:-mean}).

\paragraph*{\emph{Atari 2600 games}}

We used OpenAI's Gym environments to simulate Atari games. We used
the standard environment settings, employing no-frame-skip versions
of the games. The picture of the game snapshot was preprocessed by
CNNs and the A3C agent was adopted from the original paper with default
hyper-parameters as in \cite{mnih2016asynchronous}. \emph{The actor/critic
network of A3C was LSTM whose output layer's working program was provided
by Neurocoder} or HyperNet. The hidden size of the LSTM was $512$
for all baselines. We list details of models and hyper-parameters
in  Table \ref{tab:param} and \ref{tab:param-1}. 

\emph{Seaquest and MsPacman}. The original A3C agent was able to learn
and obtain a moderate score of around $2,500$ after $32$ million
environment steps. We also equipped A3C with HyperNet-based actors/critics,
however, the performance remained unchanged, with scores of about
$2/3$ of Neurocoder-based agent's.

\subsubsection*{Continual learning experiments}

\paragraph*{\emph{Split MNIST}}

We used the same $2$-layer MLP and continual learning baselines as
in \cite{hsu2018re}. Here, we again excluded program integration
to avoid catastrophic forgetting happening on the residual program
$\mathbf{R}$. We only tuned the hyper-parameters of NSM and Neurocoder
for this task. Remarkably, the NSM with much more parameters could
not improve MLP's performance, illustrating that simple conditional
computation is not enough for continual learning (see Table \ref{tab:Incremental-task-continual}). 

\paragraph*{\emph{Split CIFAR}}

The $18$-layer \emph{ResNet }implementation was adopted from Pytorch's
official release whose weights was pretrained with ImageNet dataset.
When performing continual learning with CIFAR images, we froze all
except for the output layers of ResNet, which was a $3$-layer MLP.
We only tuned the hyper-parameters of SI and Neurocoder for this task.
Details of model architectures and hyper-parameters are reported in
 Table \ref{tab:param} and \ref{tab:param-1}. 

\begin{figure}
\begin{centering}
\includegraphics[width=1\textwidth]{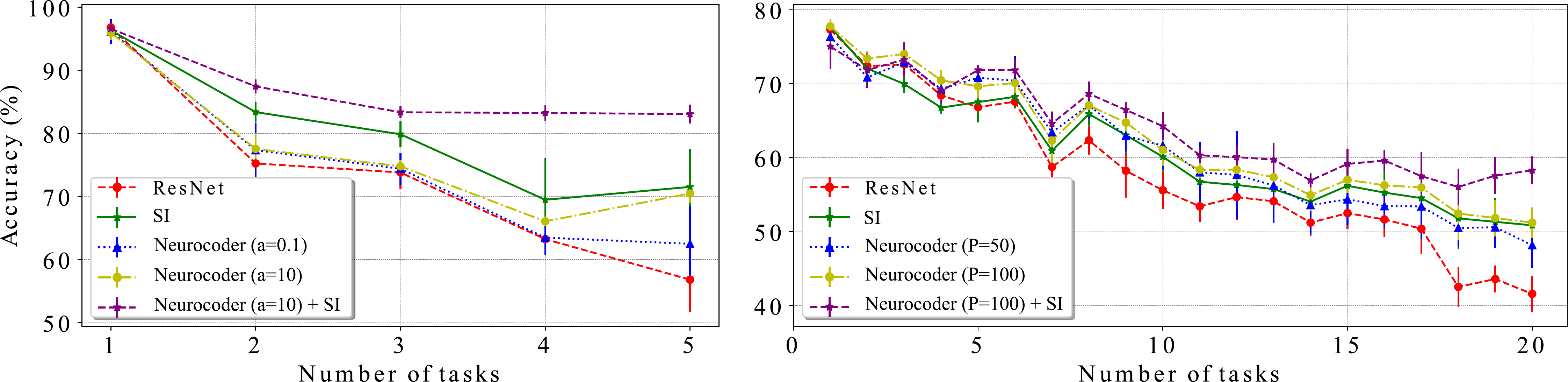}
\par\end{centering}
\caption{Incremental task continual learning with Split CIFAR10 (left) and
CIFAR100 (right). Average classification accuracy with error bar over
all learned tasks as a function of number of tasks. \label{fig:Incremental-task-continual}}
\end{figure}

In the CIFAR10 task, compared to the monolithic ResNet, the Neurocoder-augmented
ResNet could achieve much higher accuracy when we finished the learning
for all $5$ tasks ($55\%$ versus $70\%$, respectively). Also, we
realised that stressing the orthogonal loss further improved the performance.
When we employed Synaptic Intelligence (SI \cite{zenke2017continual}),
the performance of ResNet improved, yet it still dropped gradually
to just above $70\%$. In contrast, the Neurocoder-augmented ResNet
with SI maintained a stable performance above $80\%$ accuracy (see
 Fig. \ref{fig:Incremental-task-continual} (left)). 

In the CIFAR100 task, Neurocoder alone with a bigger program memory
slightly exceeded the performance of SI, which was about $10\%$ better
than ResNet. Moreover, Neurocoder plus SI outperformed using only
SI by another $10\%$ of accuracy as the number of seen tasks grew
to $20$ (see  Fig. \ref{fig:Incremental-task-continual} (right)). 

\begin{table}
\begin{centering}
\begin{tabular}{clcc}
\hline 
\multirow{3}{*}{Notation} & \multirow{3}{*}{~~~~~~~~~~~~~~Meaning} & \multicolumn{2}{c}{Location}\tabularnewline
 &  & Program  & Program\tabularnewline
 &  & Controller & Memory\tabularnewline
\hline 
 & \multicolumn{1}{l}{~~~~~~~~~~~~~~~~~~~Trainable parameters} &  & \tabularnewline
\hline 
$\theta^{u,v,\sigma}$ & Composition network & $\checkmark$ & \tabularnewline
$\phi$ & \multicolumn{1}{l}{Integration network} & $\checkmark$ & \tabularnewline
$\varphi^{u,v,\sigma}$ & \multicolumn{1}{l}{Key generator network} &  & $\checkmark$\tabularnewline
$\mathbf{R}$ & Residual program (optional) & $\checkmark$ & \tabularnewline
$\mathbf{M}_{U}$ & Memory of left singular vectors  &  & $\checkmark$\tabularnewline
$\mathbf{M}_{V}$ & Memory of right singular vectors  &  & $\checkmark$\tabularnewline
$\mathbf{M}_{S}$ & Memory of singular values  &  & $\checkmark$\tabularnewline
\hline 
 & ~~~~~~~~~~~~~~~~~~~~~Control variables &  & \tabularnewline
\hline 
$\boldsymbol{\xi}_{t}^{p}$ & Composition control signal & $\checkmark$ & \tabularnewline
$\boldsymbol{\lambda}_{t}^{p}$ & Integration control signal & $\checkmark$ & \tabularnewline
$k^{u,v,\sigma}$ & Program keys &  & $\checkmark$\tabularnewline
$m^{u,v,\sigma}$ & Program usages &  & $\checkmark$\tabularnewline
\hline 
 & ~~~~~~~~~~~~~~~~~~~~~~Hyper-parameters &  & \tabularnewline
\hline 
$P$ & Number of memory slots &  & $\checkmark$\tabularnewline
$K$ & Key dimension &  & $\checkmark$\tabularnewline
$l_{I}$ & Number of considered least-used slots &  & $\checkmark$\tabularnewline
$J$ & Number of recurrent attention steps & $\checkmark$ & \tabularnewline
$H$ & Number of attention heads & $\checkmark$ & \tabularnewline
$a$ & Orthogonal loss weight &  & $\checkmark$\tabularnewline
\hline 
\end{tabular}
\par\end{centering}
\begin{centering}
~
\par\end{centering}
\caption{Important parameters of Neurocoder.\label{tab:Neurocoder-parameters}}
\end{table}
\begin{table}
\begin{centering}
\begin{tabular}{cccccc}
\hline 
Architecture & Task & Original & MOE & NSM & Neurocoder\tabularnewline
\hline 
\multirow{2}{*}{MLP} & CIFAR10 & 52.06 & 50.76 & 52.76 & \textbf{54.86}\tabularnewline
 & CIFAR100 & 23.31 & 22.79 & 25.65 & \textbf{26.24}\tabularnewline
\hline 
\multirow{2}{*}{LeNet} & CIFAR10 & 75.71 & 75.88 & 75.45 & \textbf{78.92}\tabularnewline
 & CIFAR100 & 42.73 & 42.47 & 43.14 & \textbf{47.21}\tabularnewline
\hline 
\multirow{2}{*}{DenseNet} & CIFAR10 & 93.61 & 80.61 & 94.24 & \textbf{95.61}\tabularnewline
 & CIFAR100 & 68.37 & 38.20 & 64.53 & \textbf{72.20}\tabularnewline
\hline 
\end{tabular}
\par\end{centering}
\begin{centering}
~
\par\end{centering}
\caption{Best test accuracy over 5 runs on image classification tasks comparing
original architecture, Mixture of Experts (MOE), Neural Stored-program
Memory (NSM) and our architecture (Neurocoder). Three architectures
of the Main Network of Neurocoder were considered: $3$-layer perceptron
(MLP), $5$-layer CNN (LeNet \cite{lecun1998gradient}) and very deep
Densely Connected Convolutional Networks (DenseNet \cite{huang2017densely}).
We employed two classical image classification datasets: CIFAR10 and
CIFAR100.\label{tab:Best-accuracy-over}}
\end{table}
\begin{figure}
\begin{centering}
\includegraphics[width=1\textwidth]{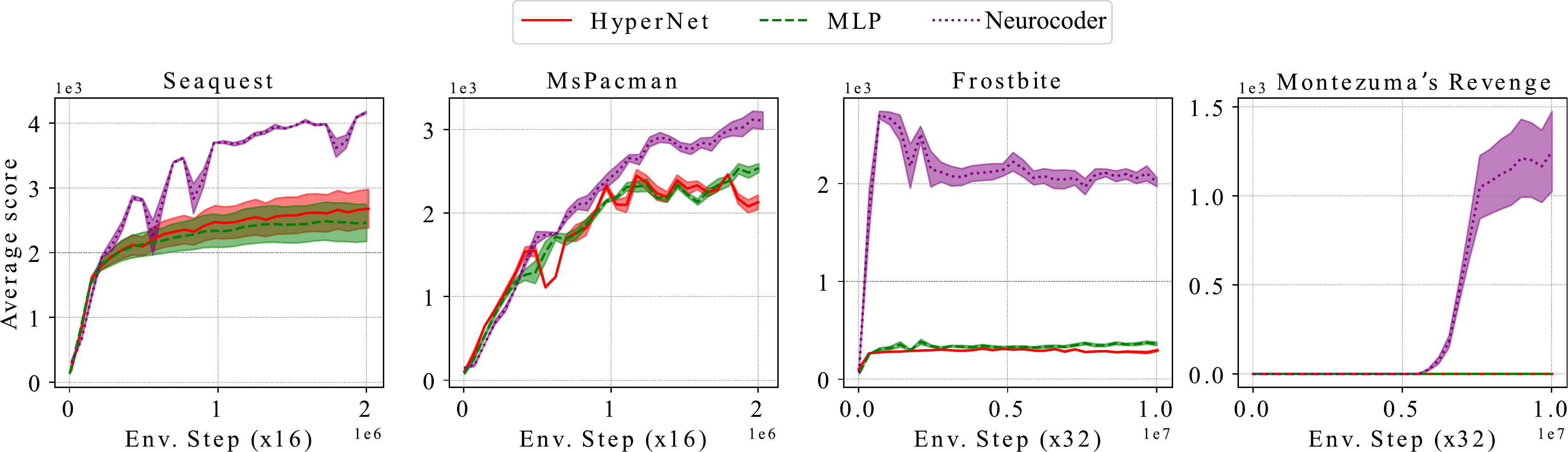}
\par\end{centering}
\caption{Learning curves (mean and std. over 5 runs) on representative Atari
2600 games. All baselines are applied to the actor/critic networks
in the A3C agent. \label{fig:atari}}
\end{figure}
\begin{table}
\begin{centering}
\begin{tabular}{ccccc}
\hline 
Method & MLP \cite{hsu2018re} & MLP (ours) & NSM & Neurocoder\tabularnewline
\hline 
Adam & 93.46$\pm$2.01 & 93.75$\pm$3.28 & 87.55$\pm$ 4.38 & \textbf{96.54$\pm$1.39}\tabularnewline
Adagrad & 98.06$\pm$0.53 & 98.02$\pm$0.89 & 96.63$\pm$1.49 & \textbf{99.01$\pm$0.19}\tabularnewline
L2 & 98.18$\pm$0.96 & 98.14$\pm$0.43  & 91.44$\pm$ 3.80 & \textbf{98.35$\pm$0.74 }\tabularnewline
SI & 98.56$\pm$0.49 & 98.69$\pm$0.20 & 98.87$\pm$0.20 & \textbf{99.14$\pm$0.24}\tabularnewline
EWC & 97.70$\pm$0.81 & 97.00$\pm$1.10 & 93.94$\pm$2.36 & \textbf{97.88$\pm$0.22}\tabularnewline
O-EWC & 98.04$\pm$1.10 & 98.23$\pm$1.17 & 96.11$\pm$1.27 & \textbf{98.30$\pm$1.48}\tabularnewline
\hline 
\end{tabular}
\par\end{centering}
\begin{centering}
~
\par\end{centering}
\caption{Incremental task continual learning with Split MNIST. Final test accuracy
(mean and std.) over 10 runs. \label{tab:Incremental-task-continual}}
\end{table}
\begin{table}
\begin{centering}
\begin{tabular}{ccccc}
\hline 
Method & MLP \cite{hsu2018re} & MLP (ours) & NSM & Neurocoder\tabularnewline
\hline 
Adam & 55.16$\pm$1.38 & 53.55$\pm$1.27 & 54.85$\pm$2.81 & \textbf{58.46$\pm$0.46} \tabularnewline
Adagrad & 58.08$\pm$1.06 & 57.83$\pm$2.74 & 58.42$\pm$1.87 & \textbf{62.28$\pm$4.03}\tabularnewline
L2 & 66.00$\pm$3.73 & 64.37$\pm$2.40 & 62.83$\pm$7.21 & \textbf{69.89$\pm$1.72 }\tabularnewline
SI & 64.76$\pm$3.09 & 64.41$\pm$3.36  & 64.36$\pm$2.99 & \textbf{67.96$\pm$3.22}\tabularnewline
EWC & 58.85$\pm$2.59 & 58.41$\pm$2.37 & 58.12$\pm$3.24 & \textbf{65.66$\pm$1.25 }\tabularnewline
O-EWC & 57.33$\pm$1.44 & 57.78$\pm$1.84 & 58.55$\pm$3.40 & \textbf{73.97$\pm$1.50}\tabularnewline
\hline 
\end{tabular}
\par\end{centering}
\begin{centering}
~
\par\end{centering}
\caption{Incremental domain continual learning with Split MNIST. Final test
accuracy (mean and std.) over 10 runs.\label{tab:Incremental-domain-continual} }
\end{table}
\begin{table}
\begin{centering}
\begin{tabular}{ccc}
\hline 
Task & \multicolumn{2}{c}{Neurocoder}\tabularnewline
\hline 
\multirow{2}{*}{MNIST} & \multicolumn{2}{c}{$P=5,J=5,H=1$}\tabularnewline
 & \multicolumn{2}{c}{$K=2,l_{I}=2,a=0.1$}\tabularnewline
\multirow{2}{*}{CIFARs} & \multicolumn{2}{c}{$P=30,J=5,H=3$}\tabularnewline
 & \multicolumn{2}{c}{$K=5,l_{I}=5,a=0.1$}\tabularnewline
Polynomial  & $P=10,J=1,H=15$ & $P=20,J=5,H=1$\tabularnewline
auto-regression & $K=3,l_{I}=2,a=0.1$ & $K=3,l_{I}=2,a=0.1$\tabularnewline
\multirow{1}{*}{Atari games} & \multicolumn{2}{c}{$P=80,J=1,H=15,K=3,l_{I}=5,a=0.1$}\tabularnewline
\multirow{1}{*}{Split MNIST} & \multicolumn{2}{c}{$P=50,J=1,H=10,K=5,l_{I}=5,a=10$}\tabularnewline
\multirow{1}{*}{Split CIFARs} & \multicolumn{2}{c}{$P=100,J=1,H=10,K=5,l_{I}=5,a=10$}\tabularnewline
\hline 
\end{tabular}
\par\end{centering}
\begin{centering}
~
\par\end{centering}
\caption{Best hyper-parameters of Neurocoder in all experiments. -- denotes
not available. For polynomial auto-regression task, two Neurocoder
configurations are tested, corresponding to single-step and multi-step
Neurocoder. Across experiments, MOE and NSM employ $10$ experts and
$10$ memory banks, respectively. HyperNet does not have any special
hyper-parameters. \label{tab:param}}
\end{table}
\begin{table}
\begin{centering}
{\footnotesize{}}%
\begin{tabular}{cccccccc}
\hline 
{\footnotesize{}Task} & {\footnotesize{}Main Network} & {\footnotesize{}Original} & {\footnotesize{}MOE} & {\footnotesize{}NSM} & {\footnotesize{}HyperNet} & \multicolumn{2}{c}{{\footnotesize{}Neurocoder}}\tabularnewline
\hline 
{\footnotesize{}MNIST} & {\footnotesize{}Linear classifier} & {\footnotesize{}7.8K} & {\footnotesize{}--} & {\footnotesize{}--} & {\footnotesize{}--} & \multicolumn{2}{c}{{\footnotesize{}7.3K}}\tabularnewline
\multirow{3}{*}{{\footnotesize{}CIFARs}} & \multirow{1}{*}{{\footnotesize{}$3$-layer MLP}} & {\footnotesize{}1.7M} & {\footnotesize{}15.4M} & {\footnotesize{}21.2M} & \multirow{1}{*}{{\footnotesize{}--}} & \multicolumn{2}{c}{{\footnotesize{}1.9M}}\tabularnewline
 & \multirow{1}{*}{{\footnotesize{}LeNet}} & {\footnotesize{}2.1M} & {\footnotesize{}12.3M} & {\footnotesize{}27.1M} & \multirow{1}{*}{{\footnotesize{}--}} & \multicolumn{2}{c}{{\footnotesize{}2.3M}}\tabularnewline
 & \multirow{1}{*}{{\footnotesize{}DenseNet}} & {\footnotesize{}7.0M} & {\footnotesize{}20.5M} & {\footnotesize{}16.7M} & \multirow{1}{*}{{\footnotesize{}--}} & \multicolumn{2}{c}{{\footnotesize{}7.3M}}\tabularnewline
{\footnotesize{}Polynomial } & \multirow{2}{*}{{\footnotesize{}GRU}} & \multirow{2}{*}{{\footnotesize{}3.4K}} & \multirow{2}{*}{{\footnotesize{}--}} & \multirow{2}{*}{{\footnotesize{}--}} & \multirow{2}{*}{{\footnotesize{}3.5K}} & \multirow{2}{*}{{\footnotesize{}3.6K }} & \multirow{2}{*}{{\footnotesize{}2.1K}}\tabularnewline
{\footnotesize{}auto-regression} &  &  &  &  &  &  & \tabularnewline
\multirow{1}{*}{{\footnotesize{}Atari games}} & \multirow{1}{*}{{\footnotesize{}LSTM}} & {\footnotesize{}3.2M} & \multirow{1}{*}{{\footnotesize{}--}} & \multirow{1}{*}{{\footnotesize{}--}} & {\footnotesize{}3.6M} & \multicolumn{2}{c}{{\footnotesize{}3.3M}}\tabularnewline
\multirow{1}{*}{{\footnotesize{}Split MNIST}} & \multirow{1}{*}{{\footnotesize{}$2$-layer MLP}} & {\footnotesize{}328K} &  & {\footnotesize{}2.3M} &  & \multicolumn{2}{c}{{\footnotesize{}348K}}\tabularnewline
\multirow{1}{*}{{\footnotesize{}Split CIFARs}} & \multirow{1}{*}{{\footnotesize{}ResNet}} & {\footnotesize{}12.6M} & \multirow{1}{*}{{\footnotesize{}--}} & \multirow{1}{*}{{\footnotesize{}--}} & \multirow{1}{*}{{\footnotesize{}--}} & \multicolumn{2}{c}{{\footnotesize{}12.6M}}\tabularnewline
\hline 
\end{tabular}{\footnotesize\par}
\par\end{centering}
\begin{centering}
~
\par\end{centering}
\caption{Number of parameters of machine learning models in all experiments.
The parameter count includes the parameter of the Main Network and
the conditional computing model. -- denotes not available. For tasks
that contain different datasets, leading to slightly different model
size, the numbers of parameters are averaged. For polynomial auto-regression
task, two Neurocoder configurations are tested, corresponding to single-step
and multi-step Neurocoder. \label{tab:param-1}}
\end{table}

%% file: main.bbl
\begin{thebibliography}{10}

\bibitem{bahdanau2014neural}
Dzmitry Bahdanau, Kyunghyun Cho, and Yoshua Bengio.
\newblock Neural machine translation by jointly learning to align and
  translate.
\newblock In {\em International Conference on Learning Representations}, 2015.

\bibitem{bellemare2013arcade}
Marc~G Bellemare, Yavar Naddaf, Joel Veness, and Michael Bowling.
\newblock The arcade learning environment: An evaluation platform for general
  agents.
\newblock {\em Journal of Artificial Intelligence Research}, 47:253--279, 2013.

\bibitem{bengio2013estimating}
Yoshua Bengio, Nicholas L{\'e}onard, and Aaron Courville.
\newblock Estimating or propagating gradients through stochastic neurons for
  conditional computation.
\newblock {\em arXiv preprint arXiv:1308.3432}, 2013.

\bibitem{cho2014learning}
Kyunghyun Cho, Bart van Merri{\"e}nboer, Caglar Gulcehre, Dzmitry Bahdanau,
  Fethi Bougares, Holger Schwenk, and Yoshua Bengio.
\newblock Learning phrase representations using {RNN} encoder{--}decoder for
  statistical machine translation.
\newblock In {\em Conference on Empirical Methods in Natural Language
  Processing ({EMNLP})}, pages 1724--1734. Association for Computational
  Linguistics, October 2014.

\bibitem{eccles1981modular}
JC~Eccles.
\newblock The modular operation of the cerebral neocortex considered as the
  material basis of mental events.
\newblock {\em Neuroscience}, 6(10):1839--1855, 1981.

\bibitem{edelman1993neural}
Gerald~M Edelman.
\newblock Neural darwinism: selection and reentrant signaling in higher brain
  function.
\newblock {\em Neuron}, 10(2):115--125, 1993.

\bibitem{edelman1978mindful}
Gerald~M Edelman and Vernon~B Mountcastle.
\newblock {\em The mindful brain: cortical organization and the group-selective
  theory of higher brain function.}
\newblock Massachusetts Inst of Technology Pr, 1978.

\bibitem{frackowiak2004human}
Richard~SJ Frackowiak.
\newblock {\em Human brain function}.
\newblock Elsevier, 2004.

\bibitem{french1999catastrophic}
Robert~M French.
\newblock Catastrophic forgetting in connectionist networks.
\newblock {\em Trends in cognitive sciences}, 3(4):128--135, 1999.

\bibitem{graves2014neural}
Alex Graves, Greg Wayne, and Ivo Danihelka.
\newblock Neural turing machines.
\newblock {\em arXiv preprint arXiv:1410.5401}, 2014.

\bibitem{graves2016hybrid}
Alex Graves, Greg Wayne, Malcolm Reynolds, Tim Harley, Ivo Danihelka, Agnieszka
  Grabska-Barwi{\'n}ska, Sergio~G{\'o}mez Colmenarejo, Edward Grefenstette,
  Tiago Ramalho, John Agapiou, et~al.
\newblock Hybrid computing using a neural network with dynamic external memory.
\newblock {\em Nature}, 538(7626):471--476, 2016.

\bibitem{ha2016hypernetworks}
David Ha, Andrew~M. Dai, and Quoc~V. Le.
\newblock Hypernetworks.
\newblock In {\em International Conference on Learning Representations}, 2017.

\bibitem{happel1994design}
Bart~LM Happel and Jacob~MJ Murre.
\newblock Design and evolution of modular neural network architectures.
\newblock {\em Neural networks}, 7(6-7):985--1004, 1994.

\bibitem{he2016deep}
Kaiming He, Xiangyu Zhang, Shaoqing Ren, and Jian Sun.
\newblock Deep residual learning for image recognition.
\newblock In {\em Proceedings of the IEEE conference on computer vision and
  pattern recognition}, pages 770--778, 2016.

\bibitem{hochreiter1997long}
Sepp Hochreiter and J{\"u}rgen Schmidhuber.
\newblock Long short-term memory.
\newblock {\em Neural computation}, 9(8):1735--1780, 1997.

\bibitem{hsu2018re}
Yen-Chang Hsu, Yen-Cheng Liu, Anita Ramasamy, and Zsolt Kira.
\newblock Re-evaluating continual learning scenarios: A categorization and case
  for strong baselines.
\newblock In {\em NeurIPS Continual learning Workshop}, 2018.

\bibitem{huang2017densely}
Gao Huang, Zhuang Liu, Laurens Van Der~Maaten, and Kilian~Q Weinberger.
\newblock Densely connected convolutional networks.
\newblock In {\em Proceedings of the IEEE conference on computer vision and
  pattern recognition}, pages 4700--4708, 2017.

\bibitem{hubel1988eye}
D.H. Hubel.
\newblock {\em Eye, Brain, and Vision}.
\newblock Scientific American Library series. Scientific American Library,
  1988.

\bibitem{jacobs1991adaptive}
Robert~A Jacobs, Michael~I Jordan, Steven~J Nowlan, and Geoffrey~E Hinton.
\newblock Adaptive mixtures of local experts.
\newblock {\em Neural computation}, 3(1):79--87, 1991.

\bibitem{krizhevsky2009learning}
Alex Krizhevsky, Geoffrey Hinton, et~al.
\newblock Learning multiple layers of features from tiny images.
\newblock 2009.

\bibitem{le2018variational}
Hung Le, Truyen Tran, Thin Nguyen, and Svetha Venkatesh.
\newblock Variational memory encoder-decoder.
\newblock In {\em Advances in Neural Information Processing Systems}, pages
  1508--1518, 2018.

\bibitem{le2018learning}
Hung Le, Truyen Tran, and Svetha Venkatesh.
\newblock Learning to remember more with less memorization.
\newblock In {\em International Conference on Learning Representations}, 2018.

\bibitem{Le2020Neural}
Hung Le, Truyen Tran, and Svetha Venkatesh.
\newblock Neural stored-program memory.
\newblock In {\em International Conference on Learning Representations}, 2020.

\bibitem{icml2020_1397}
Hung Le, Truyen Tran, and Svetha Venkatesh.
\newblock Self-attentive associative memory.
\newblock In {\em Proceedings of Machine Learning and Systems 2020}, pages
  2363--2372. 2020.

\bibitem{lecun1998gradient}
Yann LeCun, L{\'e}on Bottou, Yoshua Bengio, and Patrick Haffner.
\newblock Gradient-based learning applied to document recognition.
\newblock {\em Proceedings of the IEEE}, 86(11):2278--2324, 1998.

\bibitem{mnih2016asynchronous}
Volodymyr Mnih, Adria~Puigdomenech Badia, Mehdi Mirza, Alex Graves, Timothy
  Lillicrap, Tim Harley, David Silver, and Koray Kavukcuoglu.
\newblock Asynchronous methods for deep reinforcement learning.
\newblock In {\em International conference on machine learning}, pages
  1928--1937, 2016.

\bibitem{rosenbaum2018routing}
Clemens Rosenbaum, Tim Klinger, and Matthew Riemer.
\newblock Routing networks: Adaptive selection of non-linear functions for
  multi-task learning.
\newblock In {\em International Conference on Learning Representations}, 2018.

\bibitem{rumelhart1986learning}
David~E Rumelhart, Geoffrey~E Hinton, and Ronald~J Williams.
\newblock Learning representations by back-propagating errors.
\newblock {\em Nature}, 323(6088):533--536, 1986.

\bibitem{santoro2016meta}
Adam Santoro, Sergey Bartunov, Matthew Botvinick, Daan Wierstra, and Timothy
  Lillicrap.
\newblock Meta-learning with memory-augmented neural networks.
\newblock In {\em International conference on machine learning}, pages
  1842--1850, 2016.

\bibitem{schmidhuber1990making}
Jiirgen Schmidhuber.
\newblock Making the world differentiable: On using self-supervised fully
  recurrent n eu al networks for dynamic reinforcement learning and planning in
  non-stationary environm nts.
\newblock 1990.

\bibitem{schmidhuber1992learning}
J{\"u}rgen Schmidhuber.
\newblock Learning to control fast-weight memories: An alternative to dynamic
  recurrent networks.
\newblock {\em Neural Computation}, 4(1):131--139, 1992.

\bibitem{shazeer2017outrageously}
Noam Shazeer, Azalia Mirhoseini, Krzysztof Maziarz, Andy Davis, Quoc~V. Le,
  Geoffrey~E. Hinton, and Jeff Dean.
\newblock Outrageously large neural networks: The sparsely-gated
  mixture-of-experts layer.
\newblock In {\em International Conference on Learning Representations}, 2017.

\bibitem{turing1936}
A.M Turing.
\newblock On computable numbers, with an application to the
  entscheidungsproblem.
\newblock In {\em Proceedings of the London Mathematical Society}, 1936.

\bibitem{vaswani2017attention}
Ashish Vaswani, Noam Shazeer, Niki Parmar, Jakob Uszkoreit, Llion Jones,
  Aidan~N Gomez, {\L}ukasz Kaiser, and Illia Polosukhin.
\newblock Attention is all you need.
\newblock In {\em Advances in neural information processing systems}, pages
  5998--6008, 2017.

\bibitem{cogprints1380}
Christoph von~der Malsburg.
\newblock The correlation theory of brain function, 1981.

\bibitem{von1993first}
John Von~Neumann.
\newblock First draft of a report on the edvac.
\newblock {\em IEEE Annals of the History of Computing}, 15(4):27--75, 1993.

\bibitem{zenke2017continual}
Friedemann Zenke, Ben Poole, and Surya Ganguli.
\newblock Continual learning through synaptic intelligence.
\newblock {\em Proceedings of machine learning research}, 70:3987, 2017.

\end{thebibliography}
